\documentclass[letterpaper, 10 pt, journal, twoside]{IEEEtran}

\usepackage{graphicx}
\usepackage{epsfig} 
\usepackage{adjustbox}
\usepackage{siunitx}
\usepackage{booktabs}
\usepackage{tabularx}
\usepackage{multicol}  
\usepackage{multirow}  
\usepackage[cmex10]{amsmath}
\usepackage{amssymb} 
\usepackage{amsmath}               
  {

  }
\usepackage{algorithm}
\usepackage{algorithmic}

\usepackage{array}
\usepackage{epstopdf,enumerate,booktabs, setspace, float,stfloats,bm, multirow, lscape, caption, subcaption, graphbox, soul, color, xcolor, hyperref}
\usepackage{url}
\usepackage{fancyhdr}
\usepackage{color}
\usepackage[noadjust]{cite}

\usepackage{bm}

\title{Unveiling Uncertainty-Aware Autonomous Cooperative Learning Based Planning Strategy}
\author{Shiyao Zhang$^{1}$,~\IEEEmembership{Member,~IEEE}, Liwei Deng$^{1}$, Shuyu Zhang$^{2}$, Weijie Yuan$^{3}$,~\IEEEmembership{Senior Member,~IEEE}, \\ and Hong Zhang$^{4}$,~\IEEEmembership{Life Fellow,~IEEE}

\thanks{Manuscript received: May 22, 2025; Revised: October 8, 2025; Accepted: September 27, 2025.}

\thanks{This paper was recommended for publication by Editor Aniket Bera upon evaluation of the Associate Editor and Reviewers’ comments. This work was supported in part by Guangdong Regional Joint Fund for Basic and Applied Basic Research Fund (No. 2024A1515110203), in part by Shenzhen Science and Technology Program (No. SGDX20240115111759002), and in part by Meituan, and in part by High level of special funds (G03034K003) from Southern University of Science and Technology, Shenzhen, China. \textit{(Corresponding authors: Shuyu Zhang and Weijie Yuan.)}}
\thanks{$^{1}$Shiyao Zhang and $^{1}$Liwei Deng are with the School of Advanced Engineering, Great Bay University, Dongguan City, China, and also with Great Bay Institute for Advanced Study (GBIAS), Dongguan City, China (e-mail: zhangshiyao@gbu.edu.cn; liweidengdavid@gmail.com). 

$^{2}$Shuyu Zhang is with Department of Land Surveying and Geo-Informatics, The Hong Kong Polytechnic University, Hong Kong SAR, China (e-mail: lionel-shuyu.zhang@connect.polyu.hk). 

$^{3}$Weijie Yuan is with the School of System Design and Intelligent Manufacturing, Southern University of Science and Technology, Shenzhen, China (e-mail: yuanwj@sustech.edu.cn). 

$^{4}$Hong Zhang is with the Shenzhen Key Laboratory of Robotics and Computer Vision, Department of Electronic and Electrical Engineering, Southern University of Science and Technology, Shenzhen, China (e-mail: hzhang@sustech.edu.cn).}

\thanks{Digital Object Identifier (DOI): see top of this page.}
}

\begin{document}

\maketitle

\begin{abstract}
In future intelligent transportation systems, autonomous cooperative planning (ACP), becomes a promising technique to increase the effectiveness and security of multi-vehicle interactions. However, multiple uncertainties cannot be fully addressed for existing ACP strategies, e.g. perception, planning, and communication uncertainties. To address these, a novel deep reinforcement learning-based autonomous cooperative planning (DRLACP) framework is proposed to tackle various uncertainties on cooperative motion planning schemes. Specifically, the soft actor-critic (SAC) with the implementation of gate recurrent units (GRUs) is adopted to learn the deterministic optimal time-varying actions with imperfect state information occurred by planning, communication, and perception uncertainties. In addition, the real-time actions of autonomous vehicles (AVs) are demonstrated via the Car Learning to Act (CARLA) simulation platform. Evaluation results show that the proposed DRLACP learns and performs cooperative planning effectively, which outperforms other baseline methods under different scenarios with imperfect AV state information.


\end{abstract}

\section{Introduction}
\IEEEPARstart{B}y facilitating communication and interaction among formerly isolated vehicles, multi-vehicle systems have the potential to significantly accelerate task completion in transportation systems, such as platoon formation and collaborative planning operations \cite{pei2023collaborative,ir.2022.13}. The ability to perform high-performance and computationally efficient Autonomous Cooperative Planning (ACP), which entails planning for a complex system with high dimensions, nonholonomic motion, and collision avoidance constraints, is crucial for the success of these systems and tasks \cite{ma2023decentralized}. 

Nevertheless, there are several uncertainties that may degrade these cooperative planning strategies. First, the model-based planning can be impacted by perception uncertainty, also referred to as errors of the learning-based perception. Furthermore, certain hostile circumstances, such as extreme weather, inevitably cause a discrepancy between the intended and actual trajectories. Additionally, even though the perception uncertainty can be greatly decreased when switching from single- to multi-vehicle perception, communication outages could occur due to imperfect channel state information, endangering the information fusion \cite{li2023edge,li2024edgeacceleratedrobotnavigation}.

The current state of uncertainty-aware planning techniques can be divided into two categories: multi-vehicle uncertainty-aware motion planning \cite{FIROOZI2021104714, 10032163, 9793623, 8911491, 9801548, 10073958, 8370703}, and single-vehicle uncertainty-aware motion planning \cite{jasontits, li2023edge}. They implement perception, planning, and communication uncertainties independently in the interim. In addition, instead of adopting the multi-vehicle ACP frameworks, they primarily concentrate on single-vehicle AD. Besides, to address the uncertainty issue with incomplete information among AVs in each vehicle platoon system, the application of reinforcement learning (RL) techniques can help reaching the optimal solutions of the motion operations with imperfect information. Since there are various privacy concerns in real-world operations, AVs that may be owned by different companies are reluctant to share privacy to other AVs. As a result, such the systems incur incomplete information. There are some vehicle platooning works using RL methods \cite{9585638,9410239,9951132,10400390,10586903,10328545}. Since they can tackle the issue caused by the motion uncertainty, none of them further consider either the perception or the vehicle-to-vehicle (V2V) communication uncertainties.

To fill this gap, this paper proposes a Deep RL-based Autonomous Cooperative Planning (DRLACP) framework, which incorporates perception and communication uncertainties into the entire problem formulation. Specifically, the proposed model considers the LiDAR sensor as an illustration for computing the perception uncertainty \cite{8936542,xu2022fast} and the communication outage model based on the wireless channel distribution \cite{li2023edge}. By implementing the collision-free model, the formulated problem is derived as a learning-based cooperative model predictive control (MPC) problem. To solve this problem, a soft actor-critic (SAC) with gate recurrent units (GRUs) is adopted to learn and perform the time-varying AV actions effectively. Eventually, the proposed model is timely interacted with the Car Learning to Act (CARLA) simulation platform \cite{dosovitskiy2017carla}. 
To the best of our knowledge, this is the first work to consider multi-uncertainty aware mechanism in deep RL-based ACP system with collision avoidance constraints. The main contributions are summarized below:
\begin{itemize}
\item We propose an effective collision-free DRLACP strategy for tackling the perception and communication uncertainties in AV motion planning tasks;
\item We design a SAC approach with the implementation of GRUs to learn the deterministic time-varying actions of participated AVs, which is suitable for time-varying stochastic driving environments;
\item We evaluate the performance of the proposed strategy in the CARLA with extensive comparisons, which demonstrates its efficient performance under different scenarios.
\end{itemize}

\section{Related Work}
\subsection{Optimization-Based Cooperative Driving}
Extensive studies have investigated the applications of cooperative AV driving schemes, there are numerous studies have investigated the effect of vehicle platoon motion planning strategies. For instance, \cite{FIROOZI2021104714} developed various autonomous navigation frameworks for congested multi-lane platoons. To cope with the V2V communication module, \cite{10032163} proposed a joint vehicle platoon control and latency minimization problem with the use of Rate-Splitting Multiple Access (RSMA). In addition, \cite{9793623} developed a distributed controller to optimize the fuel consumption of a vehicle platoon via the efficient V2V communication protocol. Similar idea was also applicable to cross-road \cite{8911491}. Besides, \cite{9801548} designed a distributed platoon control approach (DPCA) in two stages, which implemented a sufficient condition for ensuring a safe junction crossing. Last but not least, the optimization problems for multi-vehicle motion planning (MVMP) were formulated to achieve the best driving actions \cite{10073958, 8370703}. However, the lack of implementing time varying perception specifications may not react well for vehicle platoon motion operations due to the neglect of real-time uncertainties.

\subsection{RL-Based Cooperative Driving}
Considering imperfect information inside the platoon system, the aforementioned studies cannot find their optimal solutions due to the uncertainty characteristic of AD. Therefore, the application of reinforcement learning (RL) techniques can help solve the games with imperfect information. \cite{9585638} proposed a hybrid Deep RL and Genetic algorithm for the platoon system, which leveraged the reduction of computational complexity and accommodation on the dynamic platoon conditions. In addition, to implement the V2V communication module, \cite{9410239} devised a communication proximity policy optimization (CommPPO) algorithm to solve the platoon control problem, which could handle various platoon dynamics. Besides, \cite{9951132} adopted an integrated DRL and dynamic programming (DP) approach to learn autonomous group control policies that embed the finite horizon value iteration framework of the deep deterministic policy gradient (DDPG) algorithm. In \cite{10400390}, a novel system framework with RL and MPC methods was proposed for AVs to perform via the routing decisions with traffic congestion criteria. \cite{ir.2022.11} adopted federated RL-based method for AV platoon control by investigating both the inter-platoon and intra-platoon environments. Furthermore, a network learning-based model predictive vehicle trajectory control structure was proposed to track time-specific velocity and position profiles \cite{10586903}. Last but not least, \cite{10328545} proposed a new LSTM-based distributed model predictive control (DMPC) ensemble method, which it developed and trained a vehicle acceleration prediction model based on a long-short-term memory (LSTM) network using real driving data. 

\subsection{Uncertainty Consideration}
Even if the above research works \cite{9585638,9410239,9951132,10400390,10586903,10328545} can deal with the motion uncertainty in cooperative platooning, they assume no sudden collisions amongst the participated AVs occur. In practice, sudden road conditions need to be accounted from on-board sensors or information shared by surrounding vehicles so as to immediately proceed the motion modifications in the system. For instance, some ego driving model, e.g. \cite{10938329,kou2025enhancinglargevisionmodel}, could help avoid perception uncertainties to control vehicle motions. 
In comparison of these studies, our work implements collision avoidance model to address the sudden road conditions, which also integrates perception and communication uncertainties into a complete cooperative driving framework.

\section{System Overview}

\begin{figure}[!t]
    \centering
    \includegraphics[width=0.49\textwidth]{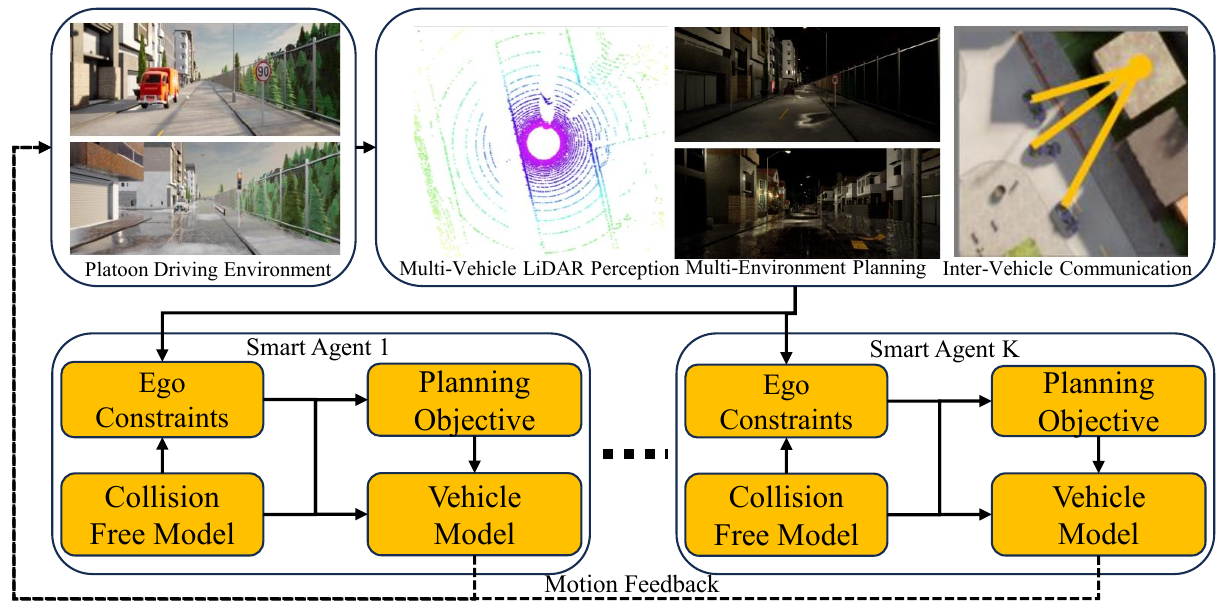}
    \caption{System architecture of the proposed DRLACP framework.}
    \label{system}
    \vspace{-0.2in}
\end{figure}

The system structure of DRLACP framework is shown in Fig.~\ref{system}, which is an integrated learning-based system with collision-free and multi-uncertainty models.  
The entire procedure of the DRLACP framework at every time step is shown as follows. First of all, platoon driving environment is generated in the system via the point-cloud based data. Then, the multi-uncertainties, including perception and communication uncertainties, are obtained from the LiDAR occlusion level, weather condition, and the wireless channel distribution, respectively. After that, the platoon operator deliver high-level reference to each smart agent (i.e. AV) so as to construct the ego constraints individually. The vehicle model at each smart agent $k$ that is incorporated the ego constraints, collision-free model, and its own planning objective, aims to solve the problem via the proposed SAC method to generate the desired trajectories and actions. Note that the objective of problem involves not only the distance deviation of the path tracking, but also the penalization for collision-free conditions among different vehicles. By checking the safety condition for pre-collision, the current and predictive driving actions are obtained to feedback the driving environment so as to prepare the next-state actions. Meanwhile, the predictive driving actions are generated with the application of GRU due to their time-series characteristics.
Last but not least, the input vehicle states are obtained in three ways: 1) onboard LiDAR sensor that directly; 2) information from other vehicles received at the onboard unit (OBU); 3) feedback controller.
The output includes collision-free trajectories and related AV actions.

\section{Methodology}
In this section, we formulate the collision-free platoon motion planning problem. The related parameters and variables are shown in Table \ref{table.P2}.

\begin{table}[t]
\small
\caption{Table of Notations for Platoon Control System.}
\centering
\label{table.P2}
\begin{tabularx}{0.95\columnwidth}{cX} 
\toprule
\multicolumn{2}{c}{Key Parameters and Control Variables} 
\\ \cmidrule(r){1-2}
$\mathcal{K}$ & Set of AV number \\
$\boldsymbol{z}_{k,t}$ & State vector of AV $k$ at time $t$\\
$x_{k,t}$ & Longitudinal position of AV $k$ at time $t$\\
$y_{k,t}$ & Lateral position of AV $k$ at time $t$\\
$\phi_{k,t}$ & Heading angle of AV $k$ at time $t$\\
$v_{k,t}$ & Velocity of AV $k$ at time $t$\\
$\mathcal{T}$ & Set of time period\\
$\Delta t$ & Time slot \\
$\boldsymbol{u}_{k,t}$ & Control input of AV $k$ at time $t$\\
$a_{k,t}$ & Acceleration of AV $k$ at time $t$\\
$\delta_{k,t}$ & Steering angle of AV $k$ at time $t$\\
$\beta_{k,t}$ & Side slip angle of AV $k$ at time $t$\\
$\mathbf{RO}$ & Orthogonal rotation matrix \\
$\textbf{t}_{r}$ & Translation vector \\
$z^{\text{Ref}}_{k,\mathcal{T}}$ & Reference trajectory of AV $k$\\
$\mathcal{P}(\boldsymbol{z}_{k,t})$ & polytope occupancy of AV $k$ at time $t$\\
$d^{\text{min}}$ & Minimum safe distance of two vehicles  \\
$W$ & Lane width \\
$\lambda^{l}_{k}$ & Length of AV $k$  \\
$\lambda^{w}_{k}$ & Width of AV $k$\\
$\boldsymbol{o}_{j}^{(k,t)}$ & Random deviation of detected object $j$ related to AV $k$ at time $t$\\
$\rho_{k}^{(j,t)}$ & Confidence score of detected object $j$ related to AV $k$ at time $t$\\
$\bm{g}_{k,t} $ & Channel coefficient of AV $k$ at time $t$\\
$\epsilon$ & Quality coefficient of the channel state information\\
$\bm{e}_{k,t-1}$ & Random variable accounting for the time-varying changes\\
$P_{j}^{(k,t)}$ & Outage probability of AV $k$ received from AV $j$ at time $t$\\
$\sigma_{t}$ & Communication outage function at time $t$\\
$d^{(j,t)}_{k}$ & Real-time distance for AV $k$ to away from AV $j$\\
\bottomrule
\end{tabularx}
\label{notations}
\end{table}

\vspace{-1mm}
\subsection{Vehicle Dynamics}  \label{sec:vehicle}
\textcolor{black}{We model each autonomous vehicle (AV) $k$ using a discrete-time kinematic formulation. Its state at time $t$ is $\boldsymbol{z}_{k,t}=[x_{k,t}, y_{k,t}, \phi_{k,t}, v_{k,t}]^{T}$, where $(x_{k,t},y_{k,t})$ denote the position and $(\phi_{k,t},v_{k,t})$ the heading and velocity (Table~\ref{table.P2}). The control input is $\boldsymbol{u}_{k,t}=[a_{k,t}, \delta_{k,t}]$, with $a_{k,t}$ and $\delta_{k,t}$ being the acceleration and steering angle. The side slip angle $\beta_{k,t}$, state update, and geometric transformations (rotation $\mathbf{RO}$ and translation $\mathbf{t}_{r}$) follow standard kinematic models \cite{FIROOZI2021104714}.}

\subsection{Vehicle Platoon Model}

In the proposed AV platoon system, considering the AV motion plans obtained by the high level planner, we denote the reference trajectory for AV $k$ as $z_{k,\mathcal{T}}^{\text{Ref}}$. \textcolor{black}{Then, the real-time state vector of AV $k$ at time $t+1$ can be}
\begin{equation}\label{eqn_6}
\boldsymbol{z}_{k,t+1}=f[\boldsymbol{z}_{k,t},\boldsymbol{u}_{k,t}].
\end{equation}

For each AV $k$, the real-time state vector $z_{k,t}$ shall follow the state limits [$\boldsymbol{z}_{k}^{\text{min}}$, $\boldsymbol{z}_{k}^{\text{max}}$]. In addition, the control input $\boldsymbol{u}_{k,t}$ follow the lower and upper bounds as [$\boldsymbol{u}_{k}^{\text{min}}$, $\boldsymbol{u}_{k}^{\text{max}}$], as well as the changes in AV $k$'s acceleration rate $a_{k,t}$ and the steering angle $\delta_{k,t}$ holding $\boldsymbol{u}_{k,t} - \boldsymbol{u}_{k,t-1} \in [\boldsymbol{\Delta u}_{k}^{\text{min}}, \boldsymbol{\Delta u}_{k}^{\text{max}}]$.

Furthermore, since other AVs in the system can be regarded as moving polytopes, we need to consider the real-time states for both the leader vehicle (LV) and follower vehicles (FVs) in the platoon system. To consider the practical road conditions, the vehicle polytope occupancy must be accounted for assisting the safe driving in order to avoid the vehicle collision for all AVs. Thus, we have  
\begin{equation}\label{eqn_10}
\mathcal{P}(\boldsymbol{z}_{k,t}) \cap \mathcal{P}(\boldsymbol{z}_{j,t})=\emptyset,\   \text{if} \ k \neq j, \forall \ k, j \in \mathcal{K}, 
\end{equation}
where $\mathcal{P}(\boldsymbol{z}^{\text{LV}}_{k,t})$ denotes the real-time vehicle moving polytope for LV in the system and $\emptyset$ is the symbol of empty set.

\subsection{Collision-Free Model}
The collision-free model is developed based on the constraint \eqref{eqn_10}. Referring to \cite{NILSSON2015124}, the safe motion models can be defined by means of the affine constraints. In particular, there are two crucial constraints, namely, the Forward Collision Avoidance Constraint (FCAC) and the Rear Collision Avoidance Constraint (RCAC). The main objective of the FCAC is to prevent collision with the preceding vehicle. In this case, we can denote such the constraint as
\begin{equation}\label{eqn_11}
\frac{\Delta x_{kj, t}}{d^{\text{min}} + \lambda^{l}_{k}} \pm \frac{\Delta y_{kj, t}}{\frac{1}{2} W + \lambda^{w}_{k}} \geq 1, \ \forall k, j \in \mathcal{K},
\end{equation}
where $W$ is the lane width of the road segment and $d^{\text{min}}$ is the predefined minimum safety distance between AV $k$ and AV $j$ (reference AV). Note that $\lambda^{l}_{k}$ and $\lambda^{w}_{k}$ is the length and width of AV $k$, respectively. The sign of the second term depends on which lane the reference AV $j$ is at, i.e., `$+$' when AV $j$ is at left lane and `$-$' when AV $j$ is at right lane. Besides FCAC, the objective of RCAC is similar in formulation and the constraint is formed as
\begin{equation}\label{eqn_12}
\frac{\Delta x_{kj, t}}{d^{\text{min}} + \lambda^{w}_{k}} \pm \frac{\Delta y_{kj, t}}{\frac{1}{2} W + \lambda^{w}_{k}} \leq -1, \ \forall k, j \in \mathcal{K}.
\end{equation}

The sign of the second term depends on which lane the reference AV $j$ is at, i.e. `$+$' when AV $j$ is at right lane and `$-$' when AV $j$ is at left lane.

\subsection{Multi-Uncertainty Model}
For the LiDAR-based perception error, it refers to the limitation of hardware and the black-box feature of deep neural networks (DNNs). Such the uncertainty can cause inaccurate $\text{dist}(\mathcal{P}_{1}, \mathcal{P}_{2})$ so as to affect the collision avoidance mechanisms through FCAC and RCAC.
In this case, by following \cite{10802006}, a stochastic set, $\boldsymbol{\mathcal{O}}_{k,t} = \{\boldsymbol{o}_{j}^{(k,t)}|j\neq k\}_{j=1}^{|\mathcal{K}|}$, is defined for the uncertainty of the detected regions that covers the vehicle objects, where the random deviation $\boldsymbol{o}_{j}^{(k,t)}\in\mathbb{R}^4$ is added to the state vector $\boldsymbol{z}_{j,t}$ of AV $j$ at time $t$ (including position $(x_{j,t},y_{j,t})$, heading angle $\phi_{j,t}$, and velocity $v_{j,t}$) of object $j$ observed from vehicle $k$ at time $t$. Hence, the constraint \eqref{eqn_10} transforms to a probabilistic constraint as 
\begin{equation}\label{eqn_13}
\mathbb{P}(\mathcal{P}(\boldsymbol{z}_{k,t}) \cap \mathcal{P}(\boldsymbol{z}_{j,t}+\boldsymbol{o}_{j}^{(k,t)})\neq\emptyset|\boldsymbol{o}_{j}^{(k,t)})\leq \epsilon, ~\ \forall j \neq k, 
\end{equation}
where $\mathbb{P}(\cdot)$ is probability function and $\epsilon$ denotes the target threshold, e.g., $0.001\%$. Besides, the distribution of $\boldsymbol{\mathcal{O}}_{k,t}$ is not known in practice. So it can be reflected by the confidence score $\rho_{k}^{(j,t)}\in[0,1]$, where $\rho_{k}^{(j,t)}\rightarrow 0$ means a high $\mathcal{P}(\boldsymbol{z}_{j,t})$ and vice versa. 

On the other hand, the communication connectivity may occur upon the motion operations due to signal loss issues. In this case, it is crucial to implement the communication outage factors into the proposed model. In this case, we focus on a downlink RSMA scheme with one LV and $K$ FVs. These $K$ vehicles have a single antenna and are dispersed at random throughout the coverage region. $\bm{g}_{k} $ represents the channel state between FV $k$ and the LV, where $\bm{g}_{k} \in \mathbb{C}^{M\times1}$. We assume that the Channel State Information at the Transmitter (CSIT) knowledge is imperfect because of the vehicle's mobility and the practical wireless communication system's delay in receiving Channel State Information (CSI) messages. The real-time channel coefficient $\bm{g}_{k,t}$ can be modeled as 
\begin{equation}\label{eqn_31}
\begin{aligned}
\bm{g}_{k,t}=\epsilon\bm{g}_{k,t-1}+\sqrt{1-\epsilon^2}\bm{e}_{k,t-1}, \quad \forall k\in \mathcal{K}.
\end{aligned}
\end{equation}

Here, $\bm{g}_{k,t}$ denotes the channel coefficients at time $t$ and $\epsilon\in[0,1]$ represents the quality of the CSI. $\bm{e}_{k,t-1}\in \mathcal{CN}(0,1)$ is a random variable accounting for the time-varying changes.

At the receiving end, every FV $k$ employs a single layer of Successive Interference Cancellation (SIC) to decode $s_c $ and $s_k $. In the meantime, the interference of all other streams is treated as Gaussian noise. Thus, referring to \cite{10032137}, the instantaneous Signal-to-Interference-plus-Noise Ratio (SINR) are provided by
\begin{equation}\label{eqn_32}
\begin{aligned}
\gamma_k =\frac{|\bm{g}_k \bm{p}|^2 \rho P_t}{\sum_{k\in \mathcal{K}}|\bm{g}_k \bm{p}_i|^2 \rho_kP_t+\xi_k^2},
\end{aligned}
\end{equation}
where $\bm{p}$ is the element in the Transmit Pre-Coding (TPC) matrix and $\rho$ denotes the power allocation coefficient. Moreover, $P_t$ represents the transmission power of the LV and $\xi_k$ denotes the Gaussian noise. 

The RSMA-based communication outage $\sigma_{t}$ is a function of outage probability $P_{j}^{(k,t)}$, whereas the function increases monotonically. Their relation can be approximately expressed as $\sigma_{t} = \sigma_{0}P_{j}^{(k,t)}$, where $\sigma_{0}$ represents the transmission time for every round of position information uploading. 

In this case, the confidence after fusion at vehicle $k$ is 
\begin{align}
    \rho_{kj,t} = \max_{l=1,\cdots,K}\sigma_{t} \rho_{lj,t}.
\end{align}

Accordingly, the deviation of box $j$ at vehicle $k$ becomes $\boldsymbol{c}_{lj,t}$ (i.e., box from vehicle $l$ due to max-score fusion), where 
\begin{align}
l = \mathop{{\textrm{arg~max}}}_{l=1,\cdots,K}\sigma_{t} \rho_{lj,t}.
\end{align}

To sum up, with multi-uncertainty case, $d^{\text{min}}$ can be modified as a dynamic safety distance with the consideration of multi-uncertainties, and the real-time distance for for AV $k$ to keep away from AV $j$ is defined by
\begin{equation}\label{eqn_33}
d^{(j,t)}_{k} = 
\begin{cases}
d^{\text{min}} + (1-\rho_{k}^{(j,t)})d^{\text{max}}_{k},  \\ 
d^{\text{min}} + \left(1- \max_{l=1,\cdots,K}\sigma_{t} \rho_{lj,t} \right)d^{\text{max}}_{k},
\end{cases}
\end{equation}
where $d^{\text{max}}_{k}$ is the maximum LiDAR-based detection error.

\subsection{Problem Formulation}
\label{sec:problem_formulation}
In the practical scenario, the AV platoon cooperative lane-change motion trajectories are determined in a real-time manner due to the stochastic traffic conditions. Such the stochastic traffic conditions may easily bring out traffic jams on several lanes. Thus, it is even realistic to develop an online lane-change motion strategies for AV $k$ in the platoon system. Considering AV $K$ as one FV in the system at time $t$, it receives the real-time traffic conditions and then schedules the instantaneous lane-change motion to operate.

Given the related constraints, the objective is to determine the lane-change motion strategy of all AVs, shown as
\begin{subequations}\label{eqn_14}
\begin{align}
& \text{minimize} \quad  \sum_{k \in \mathcal{K}}\big(\sum_{s=t}^{t+T}\boldsymbol{Q}_{\boldsymbol{z}}(\boldsymbol{z}_{k, s}-\boldsymbol{z}^{\text{Ref}}_{k, s})^{2} \nonumber \\ 
& \qquad \qquad \quad + \sum_{s=t}^{t+T-1}\boldsymbol{Q}_{\boldsymbol{u}}(\boldsymbol{u}_{k, s})^{2}  + \boldsymbol{Q}_{\boldsymbol{\Delta u}}(\boldsymbol{\Delta u}_{k, t})^{2}\big),  \tag{11}
\end{align}
\end{subequations}
where $\boldsymbol{Q}_{\boldsymbol{z}}$, $\boldsymbol{Q}_{\boldsymbol{u}}$, and $\boldsymbol{Q}_{\boldsymbol{\Delta u}}$ are the weighted positive semidefinite matrices. 

The problem is apparently non-convex based on the constraints related to the kinetic model. In order to solve this problem, the use of reinforcement learning method can help get close to the optimal solutions of the original problem. Hence, the real-time reward function can be derived from the optimization problem \eqref{eqn_14}. In this case, it can be transformed to the following reward function as
\begin{equation}\label{eqn_15}
\begin{aligned}
    \boldsymbol{R}_{k,t} = & - \boldsymbol{F}_{k, t}
    - \sigma_{1} \text{max}\bigg(1 - \frac{\Delta x_{kj, t}}{d^{\text{min}} + \lambda^{l}_{k}} \mp \frac{\Delta y_{kj, t}}{\frac{1}{2} W + \lambda^{w}_{k}}, 0 \bigg) \\
    & - \sigma_{2} \text{max}\bigg(\frac{\Delta x_{kj, t}}{d^{\text{min}} + \lambda^{w}_{k}} \pm \frac{\Delta y_{kj, t}}{\frac{1}{2} W + \lambda^{w}_{k}} + 1, 0 \bigg),
\end{aligned}
\end{equation}
where $\sigma_{1}$ penalizes the reward by considering the FCAC based on \eqref{eqn_11}, while $\sigma_{2}$ penalizes the reward by implementing the RCAC based on \eqref{eqn_12}. 

\vspace{-1mm}
\section{GRU-Enhanced Soft Actor-Critic}

To address the non-convex and time-dependent nature of the AV lane-change planning problem, we propose an extended actor-critical reinforcement learning framework, termed GRU-SAC, which extends the SAC algorithm by incorporating GRUs into both policy and value networks. While SAC is an off-policy algorithm known for its sampling efficiency and stability, its original formulation assumes Markovian state transitions, which limits its ability to model temporal dependencies in sequential control tasks. In contrast, GRU-SAC explicitly captures these dependencies, allowing for more robust decision making under dynamic traffic conditions.

\subsection{Architecture of GRU-SAC}

\begin{figure}[h]
  \centering
  \includegraphics[width=0.49\textwidth]{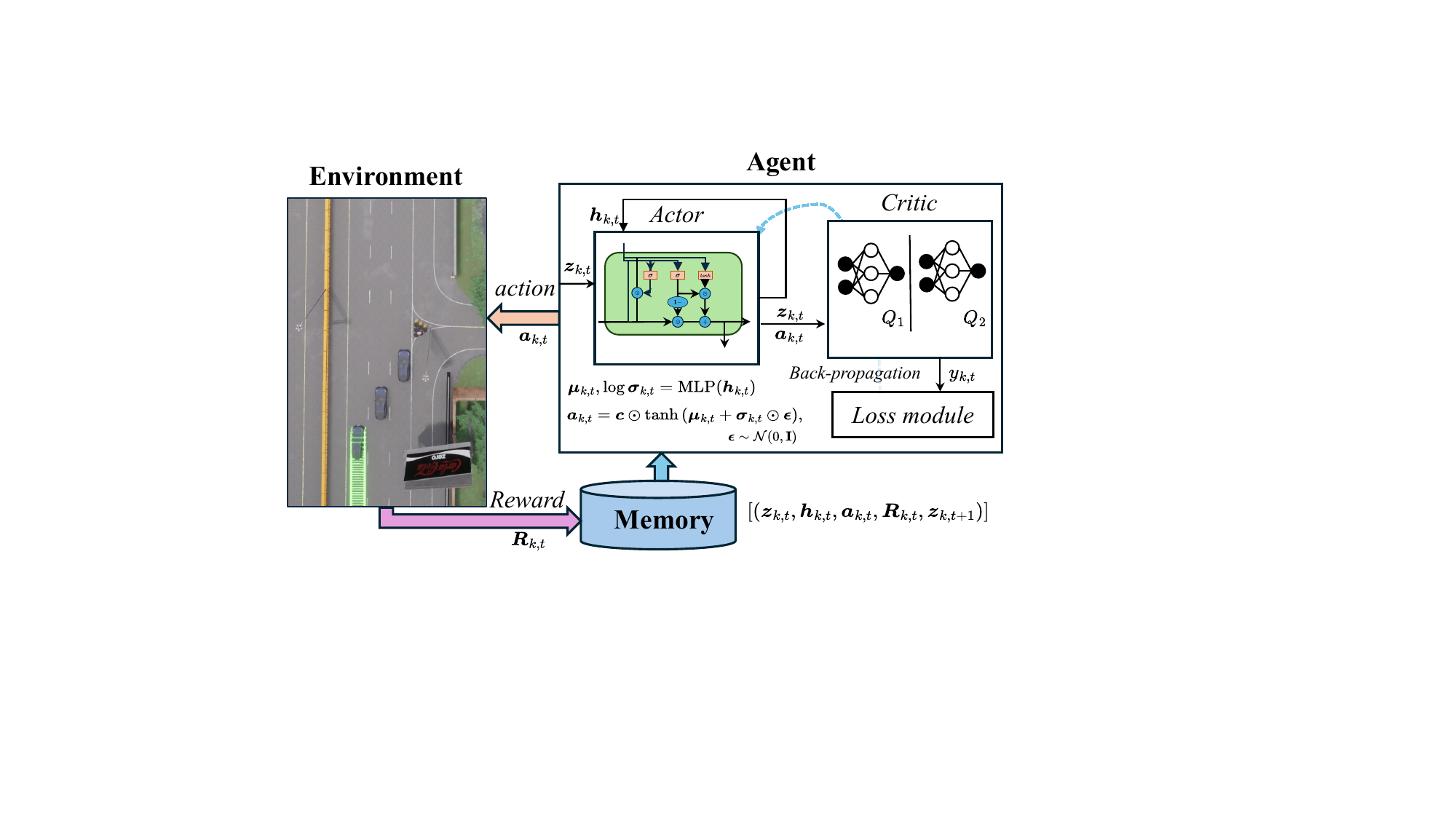}
  \caption{Overall processing flow for agent $k$ at time step $t$.}
  \label{model}
\end{figure}

Fig.~\ref{model} illustrates the complete processing flow for each agent, including the sequential state and action updates and decision-making steps. Within this framework, the GRU-SAC module, which consists of a GRU based actor and two critic networks integrated into the SAC learning loop, introduces temporal modeling into both policy generation and value estimation, enabling the agent to reason over historical observations and produce collision free driving actions.

For agent $k$ at time step $t$, the current observation $\boldsymbol{z}_{k,t}$ and the previous hidden state $\boldsymbol{h}_{k, t-1}$ are processed by a GRU to produce the updated hidden state $\boldsymbol{h}_{k,t}$. This recurrent representation captures temporal dependencies and is fed into the actor network to generate the stochastic policy $\pi(\boldsymbol{a}_{k,t}|\boldsymbol{z}_{k,t}, \boldsymbol{h}_{k,t})$, from which actions are sampled via a Gaussian distribution and squashing function. The state–action pair $(\boldsymbol{z}_{k,t},\boldsymbol{a}_{k,t})$ is simultaneously input to two independent critic networks to estimate $Q_1$ and $Q_2$, enabling clipped double Q-learning. Rewards and next observations are stored in a replay buffer, and during training, both the policy and value networks are updated with backpropagation through time. This process allows each agent to reason over sequential context, improving decision-making under dynamic and partially observable conditions.

\vspace{-1mm}
\subsection{Network Details}

The GRU-SAC framework contains three key neural components: a GRU-based actor network and two parallel GRU-based critic networks. All networks are designed with modular architectures that combine recurrent and feedforward layers to support sequential decision-making.

\paragraph{Actor Network}
The actor receives the current state $\boldsymbol{z}_{k,t}$ and the previous hidden state $\boldsymbol{h}_{k,t-1}$ as input to a GRU cell, which outputs an updated hidden state $\boldsymbol{h}_{k,t}$. This temporal representation is processed by two fully connected layers with ReLU activation to generate the mean $\boldsymbol{\mu}_{k,t}$ and log standard deviation $\log \boldsymbol{\sigma}_{k,t}$ of a Gaussian policy\textcolor{black}{, shown as}
\begin{equation}
\pi(\boldsymbol{a}_{k,t}|\boldsymbol{z}_{k,t}, \boldsymbol{h}_{k,t}) = \mathcal{N}(\boldsymbol{\mu}_{k,t}, \text{diag}(\boldsymbol{\sigma}_{k,t}^2)).
\end{equation}

The action is sampled using the reparameterization trick and squashed via a $\tanh$ function to enforce action bounds \textcolor{black}{by considering}
\begin{equation}
\boldsymbol{a}_{k,t} = \tanh\left(\boldsymbol{\mu}_{k,t} + \boldsymbol{\sigma}_{k,t} \odot \boldsymbol{\epsilon} \right), \quad \boldsymbol{\epsilon} \sim \mathcal{N}(0, \mathbf{I}).
\end{equation}

\paragraph{\textcolor{black}{Double-Critic Network}}
Two Q-networks, $Q_1$ and $Q_2$, are constructed independently using identical architectures. Each critic takes as input the concatenated state-action pair $[\boldsymbol{z}_{k,t}, \boldsymbol{a}_{k,t}]$ and processes it through a dedicated GRU cell to capture historical dependencies in the state-action dynamics. The GRU output is then passed through a stack of fully connected layers with nonlinear activations to produce a scalar Q-value as
\begin{equation}
Q_i(\boldsymbol{z}_{k,t}, \boldsymbol{a}_{k,t}) = f_{\theta_i}(\text{GRU}([\boldsymbol{z}_{k,t}, \boldsymbol{a}_{k,t}])),
\end{equation}
where $f_{\theta_i}(\cdot)$ denotes the feedforward Q-head of the $i$-th critic.

\subsection{Reward Integration}
The reward function used in GRU-SAC is derived from the trajectory optimization objective introduced in Section~\ref{sec:problem_formulation}, reflecting both control efficiency and safety considerations. At each timestep, the agent receives a scalar reward $\boldsymbol{R}_{k,t}$ that penalizes deviation from reference trajectories, excessive control effort, and potential collision risks with nearby vehicles. The reward is formulated as the negative of a weighted cost function that integrates trajectory tracking error, control smoothness, and collision avoidance constraints.

This reward directly influences the Q-value targets during critic training. Specifically, it is incorporated into the soft Bellman backup as
\begin{equation}
  \begin{aligned}
  y_{k,t} =\; & \boldsymbol{R}_{k,t} + \gamma \, \mathbb{E}_{\boldsymbol{a}_{k,t+1} \sim \pi} \Big[ \min \left(Q_1', Q_2'\right) \\
  & - \alpha \log \pi(\boldsymbol{a}_{k,t+1} \mid \boldsymbol{z}_{k,t+1}, \boldsymbol{h}_{k,t+1}) \Big],
  \end{aligned}
\end{equation}
where $\alpha$ is the temperature coefficient that controls the trade-off between reward maximization and policy entropy. By integrating the reward with temporal modeling via GRU, the agent learns policies that are not only optimal in terms of immediate feedback but also consistent over time, thereby improving decision quality in dynamic lane-change scenarios.

\subsection{Training Procedure}

The entire training process of the GRU-SAC algorithm is outlined in Algorithm~\ref{alg:gru_sac}. It includes recurrent state updates, off-policy experience sampling, soft value backups, and policy entropy regularization. At each iteration, the agent interacts with the environment and stores the observed transitions $(\boldsymbol{z}_{k,t}, \boldsymbol{a}_{k,t}, \boldsymbol{R}_{k,t}, \boldsymbol{z}_{k,t+1})$ into a replay buffer.

During training, a mini-batch of $N_\tau$ samples is randomly drawn from the buffer for gradient updates. The sampled transitions are processed in batch mode: critic networks are updated by minimizing the mean squared error between their predictions and soft Bellman targets, while the actor is updated to maximize the expected Q-value minus the entropy term. Hidden states of all GRU modules are reset at the beginning of each training step to ensure consistency in mini-batch training. Target networks are softly updated using Polyak averaging to ensure stability.

\begin{algorithm}[h]
  \caption{GRU-SAC Training Procedure}
  \label{alg:gru_sac}
  \begin{algorithmic}[1]
  \STATE Initialize network parameters $\theta_{\pi}$, $\theta_{Q_1}$, $\theta_{Q_2}$, and their target networks $\theta_{\pi'}$, $\theta_{Q_1'}$, $\theta_{Q_2'}$.
  \STATE Initialize replay buffer $\mathcal{D}$.
  \FOR{each iteration $z = 1$ to $Z$}
      \STATE Reset environment and initialize GRU hidden states.
      \FOR{$t = 1$ to $|\mathcal{T}|$}
          \STATE Observe current state $\boldsymbol{z}_{k,t}$.
          \STATE Select action $\boldsymbol{a}_{k,t} \sim \pi(\cdot|\boldsymbol{z}_{k,t}, \boldsymbol{h}_{k,t-1})$ using current policy network.
          \STATE Execute action $\boldsymbol{a}_{k,t}$, observe next state $\boldsymbol{z}_{k,t+1}$ and reward $\boldsymbol{R}_{k,t}$.
          \STATE Update GRU hidden state $\boldsymbol{h}_{k,t}$.
          \STATE Store transition $(\boldsymbol{z}_{k,t}, \boldsymbol{a}_{k,t}, \boldsymbol{R}_{k,t}, \boldsymbol{z}_{k,t+1})$ in replay buffer $\mathcal{D}$.
          \IF{training interval reached}
              \STATE Sample a mini-batch of $N_\tau$ transitions from $\mathcal{D}$.
              \STATE Compute target value $y_{k,t}$.
              \STATE Update $Q_1$ and $Q_2$ by minimizing the mean squared error to $y_{k,t}$.
              \STATE Update actor policy by minimizing:
                  $\mathcal{L}_{\pi} = \mathbb{E} \left[ \alpha \log \pi(\boldsymbol{a}_{k,t}) - \min(Q_1, Q_2)(\boldsymbol{z}_{k,t}, \boldsymbol{a}_{k,t}) \right]$.
              \STATE Softly update target networks: $\theta' \leftarrow \tau \theta + (1 - \tau)\theta'$.
          \ENDIF
      \ENDFOR
  \ENDFOR
  \end{algorithmic}
  \end{algorithm}

\section{Evaluation Results}
We first introduce the simulation setup, cooperative driving scenarios, and evaluation metrics. We then compare the proposed DRLACP (GRU-SAC) with its variants and state-of-the-art methods. Next, we present comparisons between GRU-SAC and MLP-SAC, focusing on prediction errors across different prediction window lengths. Finally, we visualize representative lane-changing cases under perception and communication uncertainties.

\subsection{Simulation Setup}
We assess the designed AV platoon system's performance in the simulations. The simulation was conducted over $T = 100$ time steps, with each time slot set to $\Delta t = 0.05$ seconds. The lane width was set to $3.7$ meters, aligned with U.S. highway regulations. Referring to \cite{DBLP:journals/corr/abs-2003-08595}, the reference trajectory $z_k^\text{Ref}$ is created by the LV. 
Considering the uncertainty condition, we consider the uniform distribution to model the uncertainty error of V2V communications. Besides, the vehicle motion of the preceding time slot $t-1$ yields the estimated trajectory $\boldsymbol{z}_{\mathcal{K},t}^\text{est}$. The upper bound of $\boldsymbol{z}_{\mathcal{K},t}$ is represented by $\boldsymbol{z}_{\mathcal{K},t}^\text{max}$. The coefficients in \eqref{eqn_14} are set as: $\boldsymbol{Q}_{\boldsymbol{z}}=[1,100,1,0.1]$ when $z_{\mathcal{K},\mathcal{T}} \in \mathbb{R}^{4 \times T}$, and $\boldsymbol{Q}_{\boldsymbol{u}}=[1,1]$ when $u_{\mathcal{K},\mathcal{T}} \in \mathbb{R}^{2 \times T}$.

The settings of each AV are presented as follows. Each AV has a length of $4.5$ meters and a width of $1.8$ meters. The changes of the AV acceleration rate changes are $-1 \text{m}/\text{s}^{2}$ and $1 \text{m}/\text{s}^{2}$, while the lower and upper bounds are set as $-4 \text{m}/\text{s}^{2}$ and $4 \text{m}/\text{s}^{2}$, respectively. Furthermore, the steering's bottom and upper bounds are set to $-0.3$ and $0.3$ radians, respectively, and its change rate is restricted to $0.2$ radian per second.

In our comparative simulations and evaluations, we implemented the proposed GRU-SAC using PyTorch and conducted simulations in Carla. 
For training the model, we set the batch size for both the actor and critic to $64$, with the actor learning rate set to $1 \times 10^{-4}$ and the critic learning rate set to $1 \times 10^{-4}$. The discount factor $\gamma$ was set to $0.99$, and the target network update rate $\tau$ was set to $0.001$. The maximum number of iterations was set to $100,000$. Additionally, we used a replay buffer of size $10,000$.
All simulations were performed on an nVidia RTX 4070 SUPER and an I9-14900.

\subsection{Assessment of AV Platoon Motion Planning}
\label{experiment1}

To validate the lane-changing capabilities of the proposed DRLACP framework, we specifically evaluate its effectiveness in coordinating a three-vehicle platoon to execute lane-change maneuvers toward a target lane under perception and communication uncertainties. Randomized simulation trials are conducted. Performance is assessed in terms of success rate, average navigation time, average velocity, average heading angle, and average computation time, whereas the proposed DRLACP is compared against four baseline methods:
\begin{itemize}
    \item \textbf{SEMPC}: A non-cooperative model predictive control (MPC) approach applied to single vehicles~\cite{jasontits};
    \item \textbf{TCMPC}: A traditional cooperative MPC scheme that assumes deterministic system dynamics~\cite{FIROOZI2021104714};
    \item \textbf{DRLACP$_{\text{MLP}}$}: An ablated version of DRLACP, wherein the GRU is replaced with a multilayer perceptron (MLP), thereby eliminating temporal feature modeling.
    \item \textbf{DRLACP$_{\text{LSTM}}$}: An ablated version of DRLACP, wherein the GRU is replaced with a LSTM network.
\end{itemize}

\begin{table*}[ht]
  \small
  \caption{Quantitative result for different methods.
  }
  \centering
  \label{table.quantitative}
  \begin{center}
  \begin{tabular}{lccccc} \toprule
  Approach & DRLACP & DRLACP$_{\text{MLP}}$ & DRLACP$_{\text{LSTM}}$ & TCMPC & SEMPC \\\midrule
  Success rate & 1.0 & 0.95 & 1 & 0.90 & 0.85 \\ 
  Navigation time (s) & 3.05 & 2.89 & 3.02 & 2.65 & 2.70 \\ 
  Averaged velocity (m/s) & 14.8401 & 14.7479 & 14.7910 & 13.7076 & 14.3510 \\ 
  Averaged heading angle (rad) & -0.1322 & -0.1221 & -0.1300 & -0.1102 & -0.1042 \\ 
  Averaged computation time (s) & 0.0066 & 0.0061 & 0.0072 & 0.12 & 0.18 \\
  \bottomrule
  \end{tabular}
  \end{center}
  \vspace{-0.2in}
\end{table*}

Table~\ref{table.quantitative} presents the quantitative performance results aggregated over 20 trials. The quantitative evaluation results demonstrate that the proposed DRLACP framework consistently outperforms the baseline methods across multiple metrics. Specifically, DRLACP achieves a $100\%$ success rate in all simulation trials, while TCMPC and SEMPC achieve $90\%$ and $85\%$ success rates, respectively. DRLACP also attains the highest average velocity, indicating efficient maneuver execution without compromising safety. Although the navigation time of DRLACP is slightly longer than that of the MPC-based approaches, it remains within an acceptable margin, reflecting a cautious and effective maneuver strategy under uncertainty. In addition, DRLACP achieves a significantly lower computation time compared to optimization-based baselines, validating its computational tractability for real-time cooperative planning applications.

The simulated findings substantiate the advantages of incorporating uncertainty modeling and temporal feature extraction within the DRLACP framework. The performance gap observed between DRLACP and its ablation variants DRLACP$_{\text{MLP}}$ and DRLACP$_{\text{LSTM}}$ highlights the critical role of temporal modeling in enhancing planning robustness and stability. The GRU- and LSTM-based DRLACP models achieve comparable performance, but the lower computational complexity of GRU gives it a slight advantage in execution time. By explicitly accounting for sequential dependencies in vehicle dynamics, DRLACP demonstrates superior cooperative behavior and reduced heading angle deviations. Overall, the results confirm that DRLACP effectively balances safety, efficiency, and real-time performance, making it a viable solution for uncertainty-aware AV platoon planning in dynamic and imperfect environments.

\subsection{Effect of Predictive Motion}

To further assess the effectiveness of employing GRU over MLP, we design a comparative simulation to evaluate prediction errors across different prediction window lengths ($H$) for GRU-SAC (DRLACP) and traditional MLP-SAC (DRLACP$_{\text{MLP}}$). The scenario involves a FV in the top lane and a tested AV required to perform a lane change to the bottom (target) lane. The setup is consistent with that described in Section~\ref{experiment1}.

\begin{figure}[h]
  \centerline{
      \includegraphics[width=0.48\textwidth]{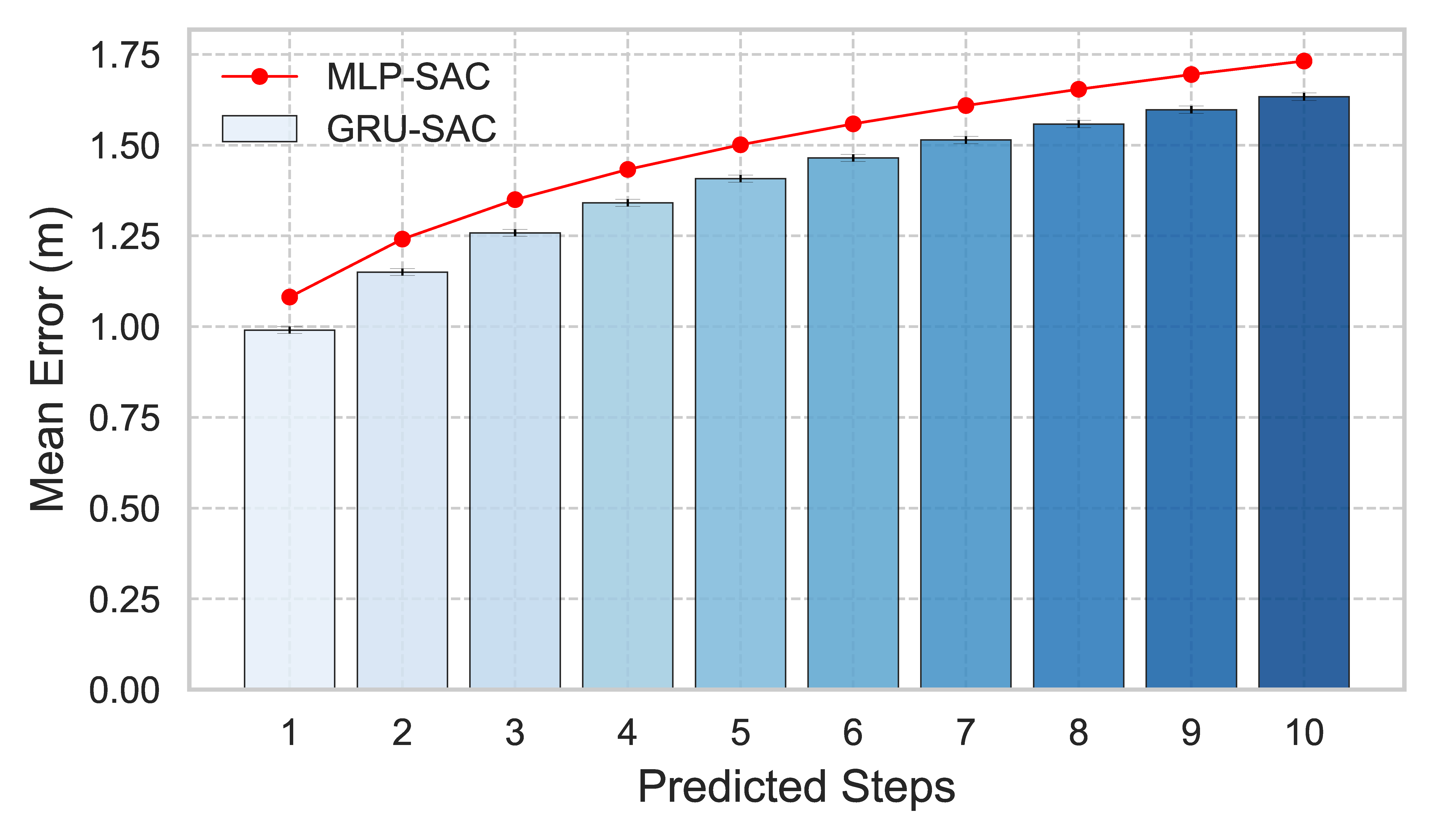}	
  }
  \caption{Predicted error versus steps via proposed GRU-SAC.}
  \label{fig:Exp2_PredstepsVsError}
  \end{figure}

In this simulation, we use an autoregressive method to predict the vehicle’s actions from the current time step $t$ to $t+H$ based on its state at $t$. The prediction performance is then evaluated by computing the Mean Absolute Error (MAE) against the reference trajectory. As shown in the Fig. \ref{fig:Exp2_PredstepsVsError}, the proposed DRLACP framework (GRU-SAC) consistently yields lower prediction errors than the MLP-SAC across all prediction window lengths. This can be attributed to the gated mechanisms in the GRU, including the reset and update gates, which enable the model to effectively capture important information from historical time steps while filtering out noise. As a result, the GRU-SAC model can trigger more accurate predictions on generated trajectories. 
In contrast, the MLP architecture processes all historical information without selective gating, which limits its ability to capture long-term dependencies and makes it more susceptible to noise, ultimately leading to degraded prediction performance.

\subsection{Evaluation Under Various Uncertainties}
\begin{figure*}[!ht]

\centering 
\begin{subfigure}{0.32\linewidth}
\vspace{1mm}
   \centering
   \includegraphics[width=1\linewidth,height=0.82\linewidth]{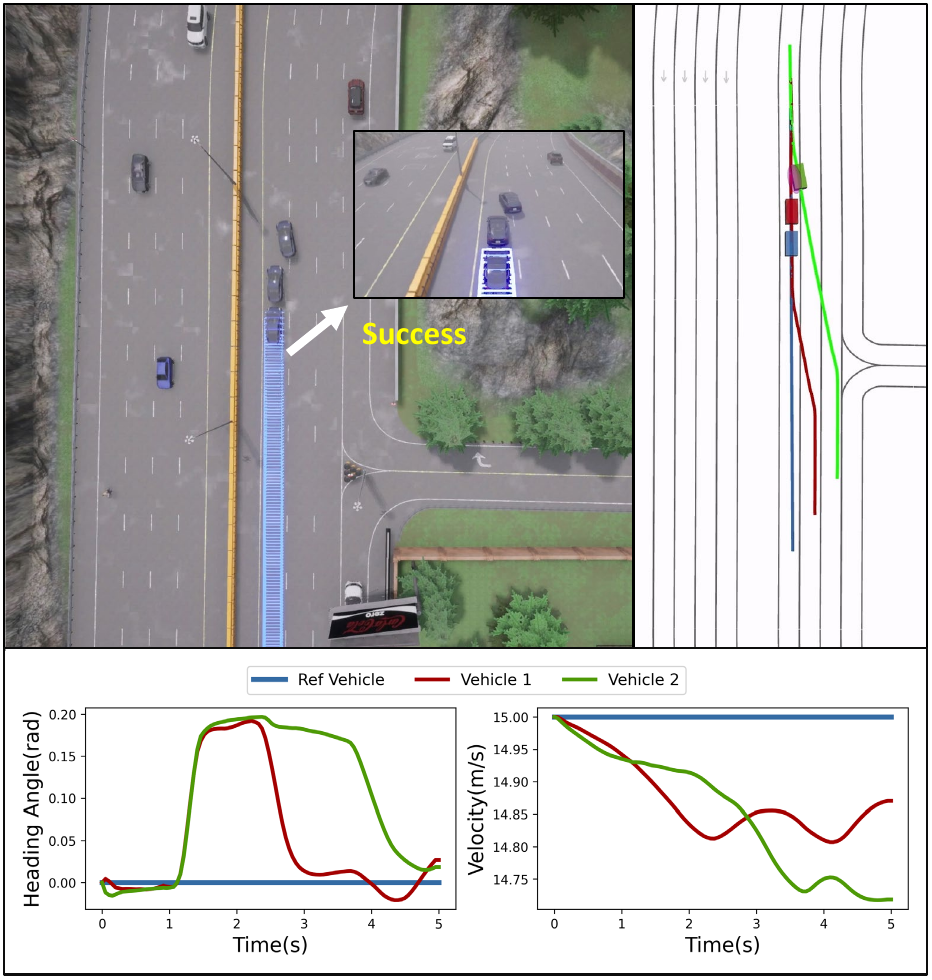}
   \caption{State and control profiles of the proposed DRLACP with 3 AVs.}
   \label{fig.2d}
\end{subfigure}
~
\centering 
\begin{subfigure}{0.32\linewidth}
\vspace{1mm}
   \centering
   \includegraphics[width=1\linewidth,height=0.82\linewidth]{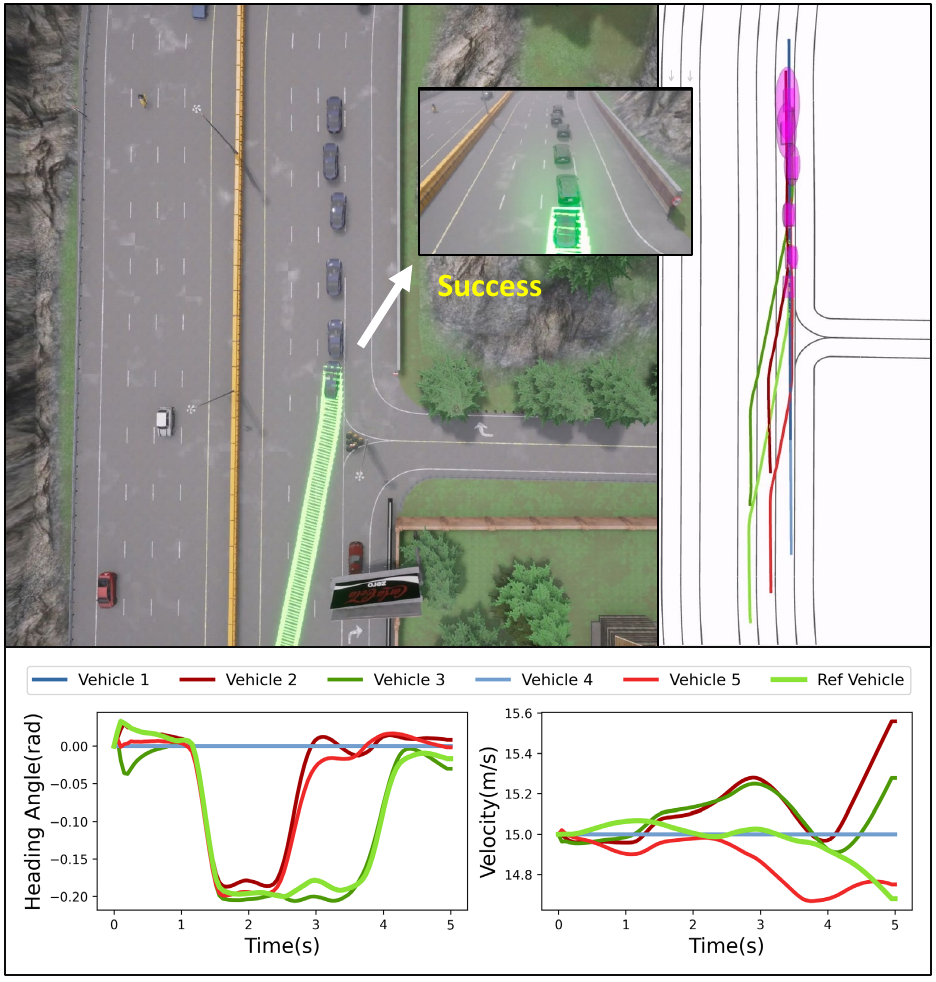}
   \caption{State and control profiles of the proposed DRLACP with 6 AVs.}
   \label{fig.2e}
\end{subfigure}
~
\centering
\begin{subfigure}{0.32\linewidth}
\vspace{1mm}
   \centering
   \includegraphics[width=1\linewidth,height=0.82\linewidth]{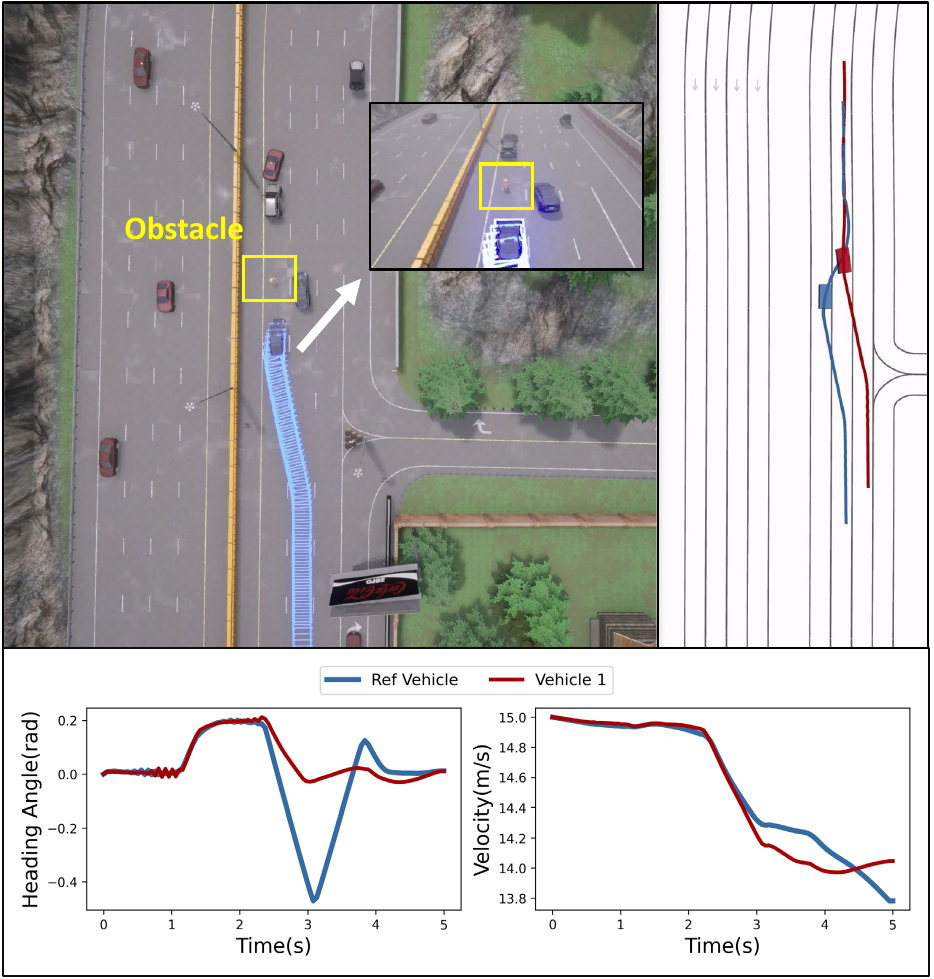}
   \caption{State and control profiles of the proposed DRLACP with 2 AVs.}
   \label{fig.2c}
\end{subfigure}

 \caption{Evaluation of the proposed DRLACP in CARLA.}
 \label{closed-loop}
\vspace{-0.1in}
\end{figure*}

To evaluate the lane-changing performance of the proposed DRLACP framework under LiDAR-based perception and communication uncertainties, we visualize the uncertainty regions of each vehicle, highlighted as purple ellipses in the upper-right corner of each subplot.
We visualize two representative scenarios in the CARLA simulation platform. In this case, the three-lane road with unique directional traffic flow occurring in a one-way traffic road environment is generated in CARLA. As shown in Fig.~\ref{fig.2d}, the proposed DRLACP framework produces smooth and collision-free trajectories while maintaining stable lane-following behavior, demonstrating its effectiveness in handling complex multi-vehicle lane-changing maneuvers. In addition, when increasing the number of participated AVs to 6, the effective and safe trajectories can still be obtained, as shown in Fig. \ref{fig.2e}. Meanwhile, these AVs' states and control profiles are shown in the bottom parts of Figs. \ref{fig.2d} and \ref{fig.2e}. The above results demonstrate the effectiveness of lane-change motions under different number of AVs in the multi-vehicle system.

Furthermore, we investigate the robustness of the proposed framework under sudden uncertainties in the same road environment. Specifically, we simulate an unexpected lane construction event occurring in the target (top) lane at $t=2.5$ seconds, when one vehicle has just completed a lane-change and another one is midway through the maneuver. In this case, due to the unexpected obstacle in the target lane, these two vehicles need to adjust its original target lane by performing collision avoidance mechanisms. Fig.~\ref{fig.2c} presents the motion planning results with timely adjustment. It is apparent that the two AVs quickly react by adjusting the target lane and successfully completing the maneuver to avoid the obstacle. These AVs' states and control profiles are shown in the bottom parts of Fig. \ref{fig.2e}. These results further confirm that the proposed DRLACP framework can operate robustly not only under regular perception and communication uncertainties but also in the presence of sudden and unexpected environmental changes.

\section{Conclusion}
This paper has proposed a novel DRLACP framework, which aims to tackle various uncertainties on cooperative motion planning schemes. In addition, this model utilizes SAC with GRUs to learn the deterministic optimal time-varying actions with imperfect vehicle state information. Evaluation results demonstrated that our proposed DRLACP performs effectively in the CARLA simulation platform and outperforms the baseline approaches via different scenarios with imperfect AV state information. What's more, the utilization of GRU can help learn and predict the continuous time-series actions in an accurate manner. Future work will consider the extension of longer-horizon temporal modeling to enhance decision-making in real-world AV applications.

\bibliographystyle{IEEEtran}
\bibliography{reference.bib}

\begin{thebibliography}{10}
\providecommand{\url}[1]{#1}
\csname url@samestyle\endcsname
\providecommand{\newblock}{\relax}
\providecommand{\bibinfo}[2]{#2}
\providecommand{\BIBentrySTDinterwordspacing}{\spaceskip=0pt\relax}
\providecommand{\BIBentryALTinterwordstretchfactor}{4}
\providecommand{\BIBentryALTinterwordspacing}{\spaceskip=\fontdimen2\font plus
\BIBentryALTinterwordstretchfactor\fontdimen3\font minus \fontdimen4\font\relax}
\providecommand{\BIBforeignlanguage}[2]{{%
\expandafter\ifx\csname l@#1\endcsname\relax
\typeout{** WARNING: IEEEtran.bst: No hyphenation pattern has been}%
\typeout{** loaded for the language `#1'. Using the pattern for}%
\typeout{** the default language instead.}%
\else
\language=\csname l@#1\endcsname
\fi
#2}}
\providecommand{\BIBdecl}{\relax}
\BIBdecl

\bibitem{pei2023collaborative}
L.~Pei, J.~Lin, Z.~Han, L.~Quan, Y.~Cao, C.~Xu, and F.~Gao, ``Collaborative planning for catching and transporting objects in unstructured environments,'' \emph{IEEE Robotics and Automation Letters}, 2023.

\bibitem{ir.2022.13}
\BIBentryALTinterwordspacing
D.~Zhu, T.~Yan, and S.~X. Yang, ``Motion planning and tracking control of unmanned underwater vehicles: technologies, challenges and prospects,'' \emph{Intelligence \& Robotics}, vol.~2, no.~3, 2022. [Online]. Available: \url{https://www.oaepublish.com/articles/ir.2022.13}
\BIBentrySTDinterwordspacing

\bibitem{ma2023decentralized}
C.~Ma, Z.~Han, T.~Zhang, J.~Wang, L.~Xu, C.~Li, C.~Xu, and F.~Gao, ``Decentralized planning for car-like robotic swarm in cluttered environments,'' in \emph{2023 IEEE/RSJ International Conference on Intelligent Robots and Systems (IROS)}.\hskip 1em plus 0.5em minus 0.4em\relax IEEE, 2023, pp. 9293--9300.

\bibitem{li2023edge}
Z.~Li, S.~Wang, S.~Zhang, M.~Wen, K.~Ye, Y.-C. Wu, and D.~W.~K. Ng, ``Edge-assisted v2x motion planning and power control under channel uncertainty,'' \emph{IEEE Transactions on Vehicular Technology}, 2023.

\bibitem{li2024edgeacceleratedrobotnavigation}
\BIBentryALTinterwordspacing
G.~Li, R.~Han, S.~Wang, F.~Gao, Y.~C. Eldar, and C.~Xu, ``Edge accelerated robot navigation with collaborative motion planning,'' 2024. [Online]. Available: \url{https://arxiv.org/abs/2311.08983}
\BIBentrySTDinterwordspacing

\bibitem{FIROOZI2021104714}
R.~Firoozi, X.~Zhang, and F.~Borrelli, ``Formation and reconfiguration of tight multi-lane platoons,'' \emph{Control Engineering Practice}, vol. 108, p. 104714, Mar. 2021.

\bibitem{10032163}
S.~Zhang, S.~Zhang, W.~Yuan, Y.~Li, and L.~Hanzo, ``Efficient rate-splitting multiple access for the internet of vehicles: Federated edge learning and latency minimization,'' \emph{IEEE Journal on Selected Areas in Communications}, vol.~41, no.~5, pp. 1468--1483, May 2023.

\bibitem{9793623}
B.~Wang and R.~Su, ``A distributed platoon control framework for connected automated vehicles in an urban traffic network,'' \emph{IEEE Transactions on Control of Network Systems}, vol.~9, no.~4, pp. 1717--1730, Dec. 2022.

\bibitem{8911491}
S.~Maiti, S.~Winter, L.~Kulik, and S.~Sarkar, ``The impact of flexible platoon formation operations,'' \emph{IEEE Transactions on Intelligent Vehicles}, vol.~5, no.~2, pp. 229--239, Jun. 2020.

\bibitem{9801548}
M.~Hu, C.~Li, Y.~Bian, H.~Zhang, Z.~Qin, and B.~Xu, ``Fuel economy-oriented vehicle platoon control using economic model predictive control,'' \emph{IEEE Transactions on Intelligent Transportation Systems}, vol.~23, no.~11, pp. 20\,836--20\,849, Nov. 2022.

\bibitem{10073958}
X.~Duan, C.~Sun, D.~Tian, J.~Zhou, and D.~Cao, ``Cooperative lane-change motion planning for connected and automated vehicle platoons in multi-lane scenarios,'' \emph{IEEE Transactions on Intelligent Transportation Systems}, vol.~24, no.~7, pp. 7073--7091, Jul. 2023.

\bibitem{8370703}
B.~Li, Y.~Zhang, Y.~Feng, Y.~Zhang, Y.~Ge, and Z.~Shao, ``Balancing computation speed and quality: A decentralized motion planning method for cooperative lane changes of connected and automated vehicles,'' \emph{IEEE Transactions on Intelligent Vehicles}, vol.~3, no.~3, pp. 340--350, Sep. 2018.

\bibitem{jasontits}
S.~Zhang, S.~Wang, S.~Yu, J.~Yu, and M.~Wen, ``Collision avoidance predictive motion planning based on integrated perception and {V2V} communication,'' \emph{IEEE Transactions on Intelligent Transportation Systems}, vol.~23, no.~7, pp. 9640--9653, July 2022.

\bibitem{9585638}
S.~B. Prathiba, G.~Raja, K.~Dev, N.~Kumar, and M.~Guizani, ``A hybrid deep reinforcement learning for autonomous vehicles smart-platooning,'' \emph{IEEE Transactions on Vehicular Technology}, vol.~70, no.~12, pp. 13\,340--13\,350, December 2021.

\bibitem{9410239}
M.~Li, Z.~Cao, and Z.~Li, ``A reinforcement learning-based vehicle platoon control strategy for reducing energy consumption in traffic oscillations,'' \emph{IEEE Transactions on Neural Networks and Learning Systems}, vol.~32, no.~12, pp. 5309--5322, December 2021.

\bibitem{9951132}
T.~Liu, L.~Lei, K.~Zheng, and K.~Zhang, ``Autonomous platoon control with integrated deep reinforcement learning and dynamic programming,'' \emph{IEEE Internet of Things Journal}, vol.~10, no.~6, pp. 5476--5489, March 2023.

\bibitem{10400390}
L.~D'Alfonso, F.~Giannini, G.~Franzè, G.~Fedele, F.~Pupo, and G.~Fortino, ``Autonomous vehicle platoons in urban road networks: A joint distributed reinforcement learning and model predictive control approach,'' \emph{IEEE/CAA Journal of Automatica Sinica}, vol.~11, no.~1, pp. 141--156, January 2024.

\bibitem{10586903}
Q.~Li, P.~Zhang, H.~Yao, Z.~Chen, and X.~Li, ``Online learning-based model predictive trajectory control for connected and autonomous vehicles: Modeling and physical tests,'' \emph{Journal of Intelligent and Connected Vehicles}, vol.~7, no.~2, pp. 86--96, June 2024.

\bibitem{10328545}
J.~Yang, D.~Chu, J.~Yin, D.~Pi, J.~Wang, and L.~Lu, ``Distributed model predictive control for heterogeneous platoon with leading human-driven vehicle acceleration prediction,'' \emph{IEEE Transactions on Intelligent Transportation Systems}, vol.~25, no.~5, pp. 3944--3959, May 2024.

\bibitem{8936542}
A.~{Eskandarian}, C.~{Wu}, and C.~{Sun}, ``Research advances and challenges of autonomous and connected ground vehicles,'' \emph{IEEE Transactions on Intelligent Transportation Systems}, vol.~22, no.~2, pp. 683--711, Feb. 2021.

\bibitem{xu2022fast}
W.~Xu, Y.~Cai, D.~He, J.~Lin, and F.~Zhang, ``Fast-lio2: Fast direct lidar-inertial odometry,'' \emph{IEEE Transactions on Robotics}, vol.~38, no.~4, pp. 2053--2073, 2022.

\bibitem{dosovitskiy2017carla}
A.~Dosovitskiy, G.~Ros, F.~Codevilla, A.~Lopez, and V.~Koltun, ``{CARLA}: {An} open urban driving simulator,'' in \emph{Proceedings of the 1st Annual Conference on Robot Learning}, ser. Proceedings of Machine Learning Research, S.~Levine, V.~Vanhoucke, and K.~Goldberg, Eds., vol.~78.\hskip 1em plus 0.5em minus 0.4em\relax PMLR, Nov. 2017, pp. 1--16.

\bibitem{ir.2022.11}
\BIBentryALTinterwordspacing
C.~Boin, L.~Lei, and S.~X. Yang, ``Avddpg - federated reinforcement learning applied to autonomous platoon control,'' \emph{Intelligence \& Robotics}, vol.~2, no.~2, 2022. [Online]. Available: \url{https://www.oaepublish.com/articles/ir.2022.11}
\BIBentrySTDinterwordspacing

\bibitem{10938329}
R.~Han, S.~Wang, S.~Wang, Z.~Zhang, J.~Chen, S.~Lin, C.~Li, C.~Xu, Y.~C. Eldar, Q.~Hao, and J.~Pan, ``Neupan: Direct point robot navigation with end-to-end model-based learning,'' \emph{IEEE Transactions on Robotics}, vol.~41, pp. 2804--2824, 2025.

\bibitem{kou2025enhancinglargevisionmodel}
\BIBentryALTinterwordspacing
W.-B. Kou, Q.~Lin, M.~Tang, J.~Lei, S.~Wang, R.~Ye, G.~Zhu, and Y.-C. Wu, ``Enhancing large vision model in street scene semantic understanding through leveraging posterior optimization trajectory,'' 2025. [Online]. Available: \url{https://arxiv.org/abs/2501.01710}
\BIBentrySTDinterwordspacing

\bibitem{NILSSON2015124}
\BIBentryALTinterwordspacing
J.~Nilsson, P.~Falcone, M.~Ali, and J.~Sjöberg, ``Receding horizon maneuver generation for automated highway driving,'' \emph{Control Engineering Practice}, vol.~41, pp. 124--133, Aug. 2015. [Online]. Available: \url{https://www.sciencedirect.com/science/article/pii/S0967066115000726}
\BIBentrySTDinterwordspacing

\bibitem{10802006}
S.~Zhang, H.~Li, S.~Zhang, S.~Wang, D.~W. Kwan~Ng, and C.~Xu, ``Multi-uncertainty aware autonomous cooperative planning,'' in \emph{2024 IEEE/RSJ International Conference on Intelligent Robots and Systems (IROS)}, 2024, pp. 1018--1025.

\bibitem{10032137}
S.~Zhang, S.~Zhang, W.~Yuan, and T.~Q.~S. Quek, ``Rate-splitting multiple access-based satellite–vehicular communication system: A noncooperative game theoretical approach,'' \emph{IEEE Open Journal of the Communications Society}, vol.~4, pp. 430--441, 2023.

\bibitem{DBLP:journals/corr/abs-2003-08595}
\BIBentryALTinterwordspacing
R.~Firoozi, X.~Zhang, and F.~Borrelli, ``Formation and reconfiguration of tight multi-lane platoons,'' \emph{CoRR}, vol. abs/2003.08595, Dec. 2020. [Online]. Available: \url{https://arxiv.org/abs/2003.08595}
\BIBentrySTDinterwordspacing

\end{thebibliography}

\end{document}